# Why Aggregate Accuracy is Inadequate for Evaluating Fairness in Law Enforcement Facial Recognition Systems

Khalid Adnan Alsayed
*Founder, Ducaltus | AI Assurance & Governance*
*BSc (Hons) Artificial Intelligence*
*School of Computing, Engineering & Digital Technologies*
Teesside University, UK
hello@ducaltus.com
s.khalid.adnan@gmail.com

**Abstract—**Facial recognition systems are increasingly deployed in law enforcement and security contexts, where algorithmic decisions can carry significant societal consequences. Despite high reported accuracy, growing evidence demonstrates that such systems often exhibit uneven performance across demographic groups, leading to disproportionate error rates and potential harm. This paper argues that aggregate accuracy is an insufficient metric for evaluating the fairness and reliability of facial recognition systems for high-stakes environments. Through analysis of subgroup-level error distribution, including false positive and false negative rates, we demonstrate how overall performance metrics can obscure critical disparities across demographic groups. Drawing on existing literature and empirical observations from classification-based systems, the paper highlights the operational risks associated with accuracy-centric evaluation practices, particularly in law enforcement applications where misclassification may result in wrongful suspicion or missed identification. We further discuss the importance of model-agnostic fairness auditing approaches that enable post-deployment evaluation without access to proprietary systems. Finally, the paper outlines the inherent trade-offs between fairness and accuracy and emphasizes the need for more comprehensive fairness-aware evaluation strategies in high-stakes AI systems.

## 1. Introduction

Facial recognition systems have become increasingly embedded within law enforcement and security infrastructures, supporting tasks such as identification, surveillance, and decision-making. Advances in deep learning and large-scale datasets have enabled these systems to achieve high levels of overall accuracy, contributing to their perception as reliable and objective tools for operational use. However, a growing body of research has demonstrated that such systems often exhibit significant performance disparities across demographic groups, particularly with respect to race and age. These disparities raise critical ethical, legal, and societal concerns, especially in high-stakes environments where algorithmic decisions may directly affect individual's rights and freedoms.

A key issue underlying these concerns is the widespread reliance on aggregate performance metrics, particularly accuracy, as the primary indicator of system reliability. While accuracy provides a general measure of overall correctness, it fails to capture how errors are distributed across different demographic subgroups. As a result, a system may appear highly accurate while disproportionately misclassifying individuals from specific populations. This limitation is particularly problematic in law enforcement contexts, where elevated false positive rates may lead to wrongful suspicion or identification, and elevated false negative rates may reduce the effectiveness of investigation processes. Empirical studies have consistently highlighted such disparities, with significantly higher error rates reported for underrepresented demographic groups in widely deployed facial recognition systems [1].

Despite increasing awareness of algorithmic bias [2], evaluation practices in real-world deployments often remain centered on aggregate metrics, with limited emphasis on subgroup-level performance analysis. This gap reflects a broader disconnect between academic research on algorithmic fairness and operational practices in applied settings. While various fairness metrics and auditing frameworks have been proposed, their adoption in practice remains inconsistent, particularly in environments where systems are proprietary or externally sourced. Consequently, there is a need to critically examine the adequacy of current evaluation paradigms and to reconsider the role of accuracy as a primary measure of system performance in high-stakes applications.

This paper argues that aggregate accuracy is insufficient for evaluating the fairness and reliability of facial recognition systems in law enforcement contexts. Drawing on existing literature and empirical insights from classification-based systems, the paper demonstrates how accuracy can obscure critical disparities in error distribution across demographic groups. It further examines the operational risks associated with the accuracy-centric evaluation, highlighting the potential for unequal outcomes and systemic bias. Finally, the paper emphasizes the importance of fairness-aware evaluation approaches that incorporate subgroup-level metrics and support more transparent, accountable deployment of artificial intelligence systems.

## 2. Limitations of Accuracy-Based Evaluation

Accuracy is one of the most widely used metrics for evaluating machine learning models, representing the proportion of correctly classified instances across a dataset. Its simplicity and interpretability have contributed to its widespread adoption in both academic research and real-world deployments. However, in the context of high-stakes applications such as law enforcement facial recognition systems, reliance on accuracy as a primary evaluation metric presents significant limitations.

A fundamental issue with accuracy is that it provides an aggregate measure of performance, offering no insight into how errors are distributed across different demographic groups. In classification tasks involving heterogeneous populations, it is possible for a model to achieve high overall accuracy while simultaneously exhibiting substantially different error rates for specific subgroups. This phenomenon occurs because accuracy treats all instances equally, without accounting for disparities in subgroup representations or the distribution of misclassification. As a result, performance inequalities that disproportionately affect certain populations may remain hidden within aggregated results.

To illustrate this limitation, consider a scenario in which a facial recognition system achieves an overall accuracy of 90%. While this figure may suggest strong performance, it does not reveal whether the system performs consistently across demographic groups. For example, one subgroup may experience a low false positive rate (FPR), while another may experience significantly higher rates of false identification. Similarly, false negative rates (FNR) may vary across groups, leading to uneven detection performance. Such disparities are not reflected in accuracy alone, highlighting its inadequacy as a standalone measure of fairness.

Figure 1 illustrates how two subgroups can exhibit identical overall accuracy while experiencing significantly different false positive and false negative rates.

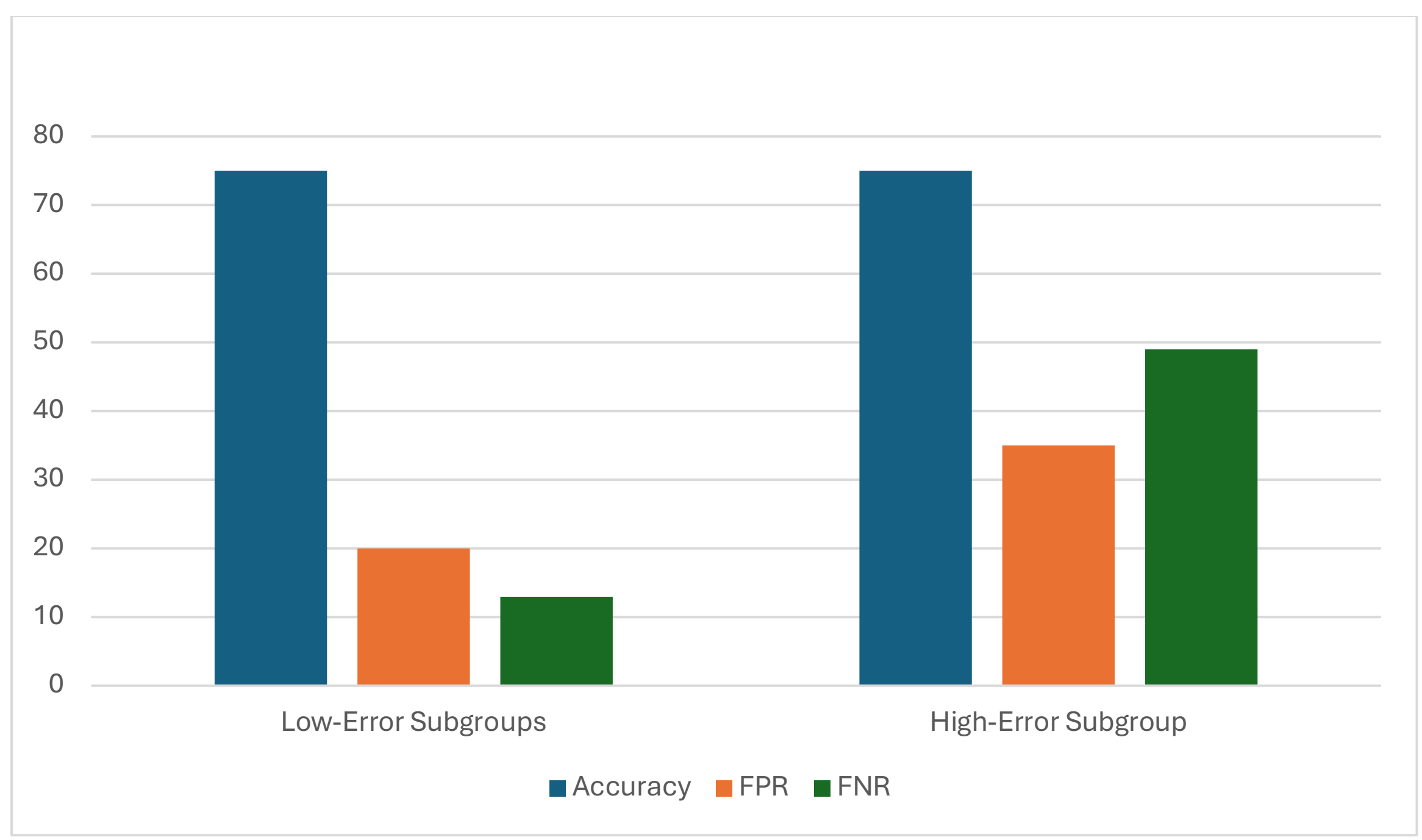

*Figure 1. Equal aggregate accuracy can conceal unequal subgroup error rates.*

This limitation is particularly critical in law enforcement contexts, where different types of errors carry distinct and potentially severe consequences. A false positive, in which an individual is incorrectly identified as a match, may lead to wrongful suspicion or investigation. Conversely, a false negative may result in failure to identify a relevant individual, potentially undermining operational objectives. Metrics such as false positive rate and false negative rate provide a more granular understanding of their error types and their distribution across demographic groups. Prior work has emphasized the importance of these metrics in fairness evaluation, particularly in domains where decision outcomes have significant societal impact [3].

Furthermore, accuracy is sensitive to class imbalance, which is common in real-world datasets. In such cases, a model may achieve high accuracy by correctly predicting the majority class while performing poorly on minority classes. When demographic groups are unevenly represented, this can exacerbate disparities in performance, further masking bias within aggregated metrics. This issue has been widely documented in studies in algorithmic bias, where models trained on imbalanced datasets exhibit reduced performance for underrepresented populations [4].

The continued reliance on accuracy as a primary evaluation metric reflects a broader challenge in aligning machine learning evaluation practices with real-world requirements. While accuracy remains useful as a general indicator of performance, it is insufficient for assessing fairness in systems that operate across diverse populations. In high-stakes domains such as law enforcement, where the cost of errors is unevenly distributed, evaluation frameworks must move beyond aggregate measures and incorporate subgroup-level analysis to ensure equitable and accountable system behavior.

## 3. Evidence of Demographic Disparities

A substantial body of research has demonstrated that facial recognition systems exhibit uneven performance across demographic groups, reinforcing concerns regarding fairness and reliability in real-world applications. One of the most widely cited studies [1] revealed that commercial facial analysis systems showed significantly higher error rates for darker-skinned individuals and women compared to light-skinned males. Their findings highlighted systemic disparities in widely deployed models, drawing attention to the limitations of relying solely on aggregate performance metrics.

Subsequent studies have confirmed that such disparities persist across both academic and industrial systems [5]. Variation in dataset composition, including imbalanced representation of demographic groups have been identified as a key contributing factor. When models are trained predominantly on data representing certain populations, they may fail to generalize effectively to underrepresented groups, resulting in higher error rates. These disparities are often not immediately visible when evaluation focuses on overall accuracy, further emphasizing the need for subgroup-level analysis.

Beyond dataset imbalance, additional factors such as variations in lighting conditions, facial features, and image quality have been shown to influence model performance differently across demographic groups. These technical challenges can interact with underlying biases in training data, compounding disparities in system outputs. As a result, even high-performing models may exhibit unequal behavior when deployed in diverse real-world environments.

Empirical observations from classification-based systems further support these findings. Analysis of subgroup-level performance metrics frequently reveals variations in both false positive rates (FPR) and false negative rates (FNR) across demographic groups, even when overall accuracy remains relatively high. This aligns with the argument presented in the previous section, where aggregate metrics obscure differences in error distribution. In practical terms, this means that two systems with similar accuracy may have significantly different fairness profiles, depending on how errors are distributed across populations.

*Table 1. Baseline overall performance metrics of the evaluated system.*

| **Metrics** | **Value** |
|---|---|
| Accuracy | 75.47% |
| False Positive Rate (FPR) | 28.42% |
| False Negative Rate (FNR) | 21.04% |

*Table 2. Range of subgroup error rates observed across demographic groups in the baseline evaluation.*

| **Demographic Attribute** | **Metric** | **Minimum Value** | **Maximum Value** |
|---|---|---|---|
| Race | FPR | 0.2015 | 0.3518 |
| Race | FNR | 0.1357 | 0.3117 |
| Age | FPR | 0.2453 | 0.4601 |
| Age | FNR | 0.1117 | 0.4958 |

These results demonstrate that error rates vary significantly across demographic groups despite a single aggregate accuracy value, reinforcing that accuracy alone is insufficient for evaluating fairness in real-world systems.

In law enforcement contexts, these disparities carry heightened significance due to the potential consequences of misclassification. Elevated false positive rates for specific demographic groups may result in disproportionate scrutiny or wrongful identification, while elevated false negative rates may reduce the effectiveness of investigative processes. These risks highlight the importance of evaluating not only whether a system is accurate, but also how that accuracy is distributed across different segments of the population.

Collectively, the existing literature and empirical observations provide strong evidence that demographic disparities are a persistent and systemic issue in facial recognition systems. These findings reinforce the argument that aggregate accuracy is insufficient as a standalone evaluation metric and underscore the need for fairness-aware approaches that explicitly account for subgroup-level performance differences.

## 4. Operational Risks in Law Enforcement

The presence of demographic bias disparities in facial recognition systems is not merely a technical concern but poses significant operational risks when such systems are deployed in law enforcement contexts. Unlike low-stakes applications, where errors may have limited consequences, misclassification in law enforcement contexts can directly affect individual's rights, freedoms, and interactions with authorities. As a result, the distribution of errors across demographic groups becomes a critical factor in assessing the real-world impact of these systems.

False positives, where an individual is incorrectly identified as a match, represent one of the most serious risks in law enforcement applications. An elevated false positive rate (FPR) for specific demographic groups can lead to disproportionate targeting, increased surveillance, or wrongful suspicion. In extreme cases, this may contribute to wrongful arrests or investigations, particularly when algorithmic outputs are treated as reliable evidence. Even when used as decision-support tools, such systems can influence human judgment, reinforce biases and increase the likelihood of unequal treatment across populations.

Conversely, false negatives, where a system fails to correctly identify an individual, also carry operational consequences. In investigation contexts, elevated false negative rates (FNR) may reduce the effectiveness of identification processes, potentially leading to missed matches or overlooked individuals of interest. While false negatives may not carry the same immediate ethical implications as false positives, they can still undermine system reliability and operational objectives. Importantly, the impact of these errors may also vary across demographic groups, further contributing to uneven system performance.

A key issue in current deployment practices is that these risks are often evaluated implicitly, rather than explicitly measured and reported. Systems are frequently assessed using aggregate performance metrics without sufficient consideration of how errors are distributed across different populations. This creates a scenario in which a system deemed "accurate" may still produce systematically unequal outcomes. In law enforcement settings, where accountability and fairness are essential, such evaluation practices are inadequate.

The operational risks associated with demographic disparities are further compounded by the opacity of many deployed systems. Facial recognition technologies are often provided by third-party vendors, limiting access to model internals, training data, and evaluation processes. This lack of transparency makes it difficult for practitioners to identify, understand, and mitigate bias in complex sociotechnical systems [6], increasing reliance on surface-level metrics such as accuracy. Without robust auditing mechanisms, these systems may be deployed with limited awareness of their potential biases and associated risks.

These challenges highlight the need to shift from purely performance-driven evaluation to risk-aware assessment frameworks. In high-stakes domains such as law enforcement, it is not sufficient for systems to be accurate on average; they must also demonstrate consistent and equitable performance across all demographic groups. Failure to account for these factors can lead to systemic bias, reduced trust in technology, and potential legal and ethical consequences. As such, understanding and addressing the operational implications of demographic disparities is essential for responsible deployment of facial recognition systems.

## 5. The Need for Fairness-Aware Evaluation

The limitations of accuracy-based evaluation and operational risks associated with demographic disparities highlight the need for more comprehensive approaches to assessing machine learning systems. In high-stakes domains such as law enforcement, evaluation frameworks must move beyond aggregate performance metrics and incorporate fairness-aware methodologies that explicitly account for differences in error distribution across demographic groups.

A key component of fairness-aware evaluation is the use of subgroup-level performance metrics. Measures such as false positive rate (FPR) and false negative rate (FNR) provide a more granular understanding of how a system behaves across different populations. Unlike overall accuracy, these metrics reveal whether certain groups are disproportionately affected by specific types of errors, enabling more informed assessment of system reliability and risk. For example, a system with balanced FPR and FNR across demographic groups is less likely to produce unequal outcomes than one where these metrics vary significantly between populations.

In addition to individual metrics, comparative measures of disparity are essential for quantifying differences in performance. One common approach is to compute the gap between the maximum and minimum values of a given metric across demographic groups [7]. This provides an interpretable measure of fairness, allowing practitioners to identify the extent of inequality in system behavior. Such disparity-based evaluations align more closely with real-world concerns, where the objective is not only to achieve high performance but also to ensure that this performance is distributed equitably.

Another important consideration is the practicality of fairness evaluation in real-world deployment scenarios. Many operational systems are proprietary or externally provided, limiting access to model internals and training data. In such cases, fairness assessment must be conducted using available outputs, such as predicted labels and associated confidence scores. Model-agnostic evaluation approaches, which operate independently of the underlying model architecture, are therefore particularly valuable. These approaches enable practitioners to audit systems post-deployment, providing insights into performance disparities without requiring modification to the original model.

Despite the availability of fairness metrics and evaluation techniques, their adoption in practice remains inconsistent. One contributing factor is the lack of accessible and interpretable tools that support fairness analysis in operational settings. Many existing frameworks are designed for research environments and require specialized expertise, limiting their applicability in real-world contexts. As a result, there is a need for evaluation approaches that are not only theoretically sound but also practical, interpretable, and deployable.

Ultimately, fairness-aware evaluation should be viewed as a fundamental component of system assessment rather than an optional extension. In high-stakes applications, it is insufficient to demonstrate that a system performs well on average, it must also be shown that the performance is consistent and equitable across all relevant populations. Incorporating subgroup-level metrics, disparity analysis, and model-agnostic auditing into evaluation practices represents a critical step towards more transparent, accountable, and responsible use of facial recognition technologies.

## 6. Trade-offs Between Fairness and Accuracy

While fairness-aware evaluation provides a more comprehensive understanding of system performance, it also introduces a fundamental challenge: the trade-off between fairness and accuracy. In many machine learning systems, particularly those operating on complex and imbalanced datasets, efforts to reduce disparities across demographic groups can lead to reductions in overall predictive performance. This tension has been widely recognized in the literature [8], [9], highlighting that achieving perfect fairness across all groups without impacting accuracy is often not feasible [8].

The underlying reason for this trade-off lies in the differing statistical distribution and representation of demographic groups within the training data, where it can achieve the greatest reduction in aggregate error. However, this optimization may come at the expense of minority or underrepresented groups, whose characteristics are less well captured by the training data. As a result, improving performance for these groups, by reducing false positive or false negative rates may require adjustments that slightly degrade the overall accuracy.

From an operational perspective, this trade-off presents a critical decision point. In high-stakes environments such as law enforcement, the cost of errors is not uniform across different types of misclassifications or across different populations. A small reduction in overall accuracy may be acceptable if it results in a substantial decrease in harmful disparities, particularly those that disproportionately affect vulnerable or underrepresented groups. Conversely, prioritizing accuracy alone may lead to systems that are efficient on average but produce inequitable outcomes in practice.

Importantly, the fairness-accuracy trade-off is not solely a technical issue but also an ethical and policy consideration. Decisions regarding acceptable levels of disparity and performance degradation must consider the broader societal contexts in which the system operates. In law enforcement applications, where decisions can have significant legal and social implications, there is a strong argument for prioritizing fairness and accountability over marginal gains in aggregate accuracy. This perspective aligns with calls for responsible AI practices that emphasise transparency, human oversight, and the mitigation of harm.

At the same time, it is essential to recognize that fairness interventions must be applied carefully and transparently. Overcorrection or poorly designed mitigation strategies may introduce new forms of bias or reduce system effectiveness beyond acceptable levels. Therefore, evaluation frameworks should not aim to eliminate trade-offs entirely but rather make them explicit and measurable. By quantifying both performance and disparity, practitioners can make informed decisions about how to balance competing objectives in each application context.

Ultimately, acknowledging and managing the trade-offs between fairness and accuracy is a key aspect of responsible AI deployment. Rather than viewing fairness as an operational constraint, it should be treated as a core component of system evaluation and design. In high-stakes domains, this requires moving beyond simplistic performance metrics and adopting evaluation approaches that reflect both technical performance and societal impact.

## 7. Conclusion

The increasing deployment of facial recognition systems in law enforcement contexts has amplified the need for robust and reliable evaluation practices. While high levels of accuracy are often presented as evidence of system effectiveness, this paper has argued that accuracy alone is insufficient for assessing fairness in high-stakes applications. Aggregate metrics fail to capture how errors are distributed across demographic groups, allowing significant disparities in performance to remain obscured.

Drawing on existing literature and empirical observations, the paper has demonstrated that facial recognition systems frequently exhibit uneven error rates across populations, particularly in terms of false positive and false negative rates. In law enforcement settings, these disparities carry substantial operational and ethical implications, including the risk of wrongful suspicion, unequal treatment, and reduced system reliability. As such, evaluation practices that rely solely on accuracy provide an incomplete and potentially misleading representation of system performance.

The analysis further highlighted the importance of fairness-aware evaluation approaches that incorporate subgroup-level metrics and disparity analysis. By examining performance across demographic groups, these methods provide a more comprehensive understanding of system behavior and enable the identification of potential biases. In addition, the discussion of model-agnostic auditing emphasises the need for practical evaluation strategies that can be applied in real-world deployment scenarios, where access to model internals is often limited.

The paper also addressed the inherent trade-off between fairness and accuracy, noting that efforts to reduce disparity may impact overall performance. In high-stakes domains, however, such trade-offs should be explicitly considered rather than ignored. Ensuring equitable system behavior and minimizing harm may, in many cases, justify modest reductions in aggregate accuracy. This perspective reinforces the need for evaluation frameworks that balance technical performance with ethical and societal considerations.

In conclusion, the findings underscore the necessity of moving beyond accuracy as the primary metric for evaluating facial recognition systems in law enforcement. Fairness-aware evaluation should be treated as a fundamental component of system assessment, enabling more transparent, accountable, and responsible deployment of artificial intelligence technologies. Future work should explore the integration of fairness auditing into operational workflows and the development of standardised evaluation protocols that better reflect the complexities of real-world applications.